\documentclass[letterpaper]{article} % DO NOT CHANGE THIS
\usepackage{aaai25}
\usepackage{times}  % DO NOT CHANGE THIS
\usepackage{helvet}  % DO NOT CHANGE THIS
\usepackage{courier}  % DO NOT CHANGE THIS
\usepackage[hyphens]{url}  % DO NOT CHANGE THIS
\usepackage{graphicx} % DO NOT CHANGE THIS
\usepackage{natbib}  % DO NOT CHANGE THIS AND DO NOT ADD ANY OPTIONS TO IT
\usepackage{caption} % DO NOT CHANGE THIS AND DO NOT ADD ANY OPTIONS TO IT
\usepackage{algorithm}
\usepackage{algorithmic}

\usepackage{newfloat}
\usepackage{listings}
\DeclareCaptionStyle{ruled}{labelfont=normalfont,labelsep=colon,strut=off} % DO NOT CHANGE THIS
\floatstyle{ruled}
\newfloat{listing}{tb}{lst}{}
\floatname{listing}{Listing}
\usepackage{amsthm}
\usepackage{amssymb}
\usepackage{amsmath}
\usepackage{mathrsfs}
\usepackage{multirow}
\usepackage{subfigure}
\usepackage{bibentry}
\nocopyright % -- Your paper will not be published if you use this command
\title{From Semantics to Hierarchy: A Hybrid Euclidean-Tangent-Hyperbolic Space Model for Temporal Knowledge Graph Reasoning}
\author{
    Siling Feng\textsuperscript{\rm 1}\equalcontrib,
    Zhisheng Qi\textsuperscript{\rm 1}\equalcontrib,
    Cong Lin\textsuperscript{\rm 2}\thanks{Corresponding author}
}
\affiliations{
    \textsuperscript{\rm 1}College of Information and Communication Engineering, Hainan University, China\\
    \textsuperscript{\rm 2}College of Electronic and Information Engineering, Guangdong Ocean University, China\\
    fengsiling@hainanu.edu.cn, charlieqi02@gmail.com, lincong@gdou.edu.cn
}

\usepackage{bibentry}
\begin{document}

\maketitle

\begin{abstract}
Temporal knowledge graphs (TKGs) have gained significant attention for their ability to extend traditional knowledge graphs with a temporal dimension, enabling dynamic representation of events over time. TKG reasoning involves extrapolation to predict future events based on historical graphs, which is challenging due to the complex semantic and hierarchical information embedded within such structured data. Existing Euclidean models capture semantic information effectively but struggle with hierarchical features. Conversely, hyperbolic models manage hierarchical features well but fail to represent complex semantics due to limitations in shallow models' parameters and the absence of proper normalization in deep models relying on the $L_2$ norm. Current solutions, such as curvature transformations, are insufficient to address these issues. In this work, a novel hybrid geometric space approach that leverages the strengths of both Euclidean and hyperbolic models is proposed. Our approach transitions from single-space to multi-space parameter modeling, effectively capturing both semantic and hierarchical information. Initially, complex semantics are captured through a fact co-occurrence and autoregressive method with normalizations in \textbf{Euclidean} space. The embeddings are then transformed into \textbf{Tangent} space using a scaling mechanism, preserving semantic information while relearning hierarchical structures through a query-candidate separated modeling approach, which are subsequently transformed into \textbf{Hyperbolic} space. Finally, a hybrid inductive bias for hierarchical and semantic learning is achieved by combining hyperbolic and Euclidean scoring functions through a learnable query-specific mixing coefficient, utilizing embeddings from hyperbolic and Euclidean spaces. Experimental results on four TKG benchmarks demonstrate that our method reduces error relatively by up to \textbf{15.0\%} in mean reciprocal rank (MRR) on YAGO compared to previous single-space models. Additionally, enriched visualization analysis validates the effectiveness of our approach, showing adaptive capabilities for datasets with varying levels of semantic and hierarchical complexity.
\end{abstract}

%    \link{Code}{https://aaai.org/example/code}
%    \link{Datasets}{https://aaai.org/example/datasets}
%    \link{Extended version}{https://aaai.org/example/extended-version}

\section{Introduction}
Knowledge graphs (KGs) are crucial in data-driven applications~\cite{zou2020survey} such as recommendation systems~\cite{guo2020survey}, medical information retrieval~\cite{yang2020biomedical}, and commonsense question-answering platforms~\cite{edge2024local}, due to their structured representation of entities and relationships~\cite{fensel2020introduction}. However, KGs often suffer from data incompleteness, driving research in KG completion~\cite{bordes2013translating,trouillon2016complex}. These approaches typically aim to represent knowledge in low-dimensional vector spaces to infer missing data. Nevertheless, the static nature of these embeddings limits their ability in modeling temporal dynamics~\cite{chang2017knowledge}.

\begin{figure}[t]
\centering
\includegraphics[width=0.95\columnwidth]{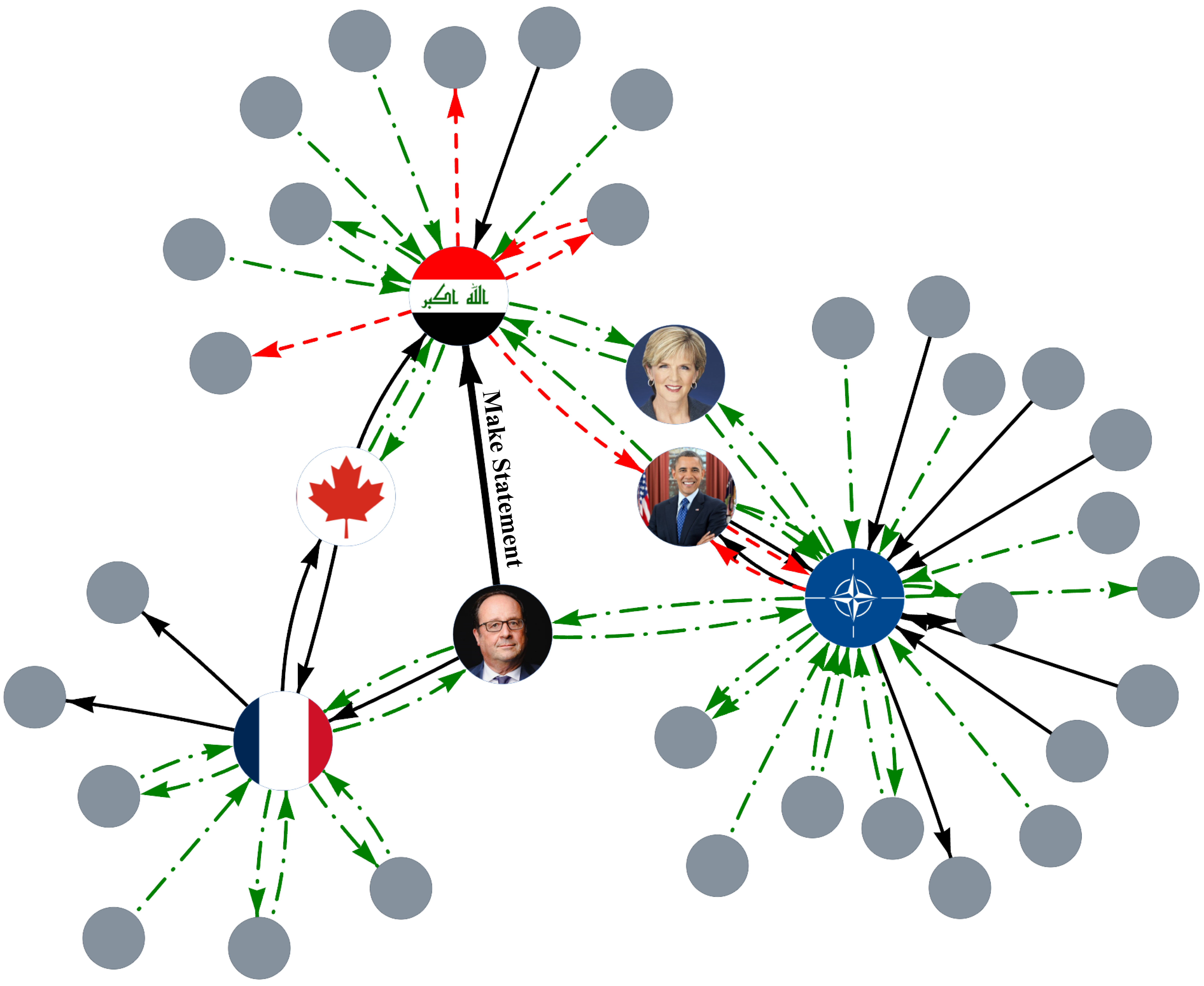}
\caption{Illustration of a TKG from ICEWS14, with line styles indicating the time of events: dot-dash for September 4th, solid for the 5th, and double-line for the 6th, all in 2014.}
\label{fig:TKG-Eg}
\end{figure}

Temporal knowledge graphs (TKGs) extend KGs by integrating a temporal dimension, transforming traditional triplets into quadruplets: (subject, relation, object, timestamp). This allows for the dynamic representation of events over time. Figure~\ref{fig:TKG-Eg} illustrates a TKG from ICEWS14~\cite{garcia2018learning}, where an event like (François Hollande, Make statement, Iraq, 2014-9-5) denotes a specific occurrence on September 5, 2014. TKG reasoning tasks generally fall into two categories: interpolation, predicting missing facts within a given time interval~\cite{dasgupta2018hyte,garcia2018learning,leblay2018deriving}, and extrapolation, forecasting future events based solely on historical data~\cite{jin2019recurrent,trivedi2017know,trivedi2019dyrep,li2021temporal}. The latter presents a significant challenge due to the absence of full context.

A deep understanding of historical data is crucial for effective extrapolation in TKGs. TKGs encapsulate both semantic and hierarchical information inherently. \textbf{Semantics} arise from graph structures and temporal dynamics, containing the intricate relationships and meanings through fact co-occurrence and sequential event information. \textbf{Hierarchy} emerges from the exponential growth of nodes, reflecting varying levels of abstraction among entities. For instance, Figure~\ref{fig:TKG-Eg} shows simultaneous and sequential events (denoted by line styles), signifying semantic connections, while the vast number of neighbors for entities like France and NATO, compared to Julie Bishop or François Hollande, suggests an underlying hierarchy.

Recent approaches \cite{jin2019recurrent, li2021temporal} integrating graph and recurrent neural networks in Euclidean space have shown promise in modeling semantic data but fall short in capturing hierarchical information. In contrast, hyperbolic geometry learning \cite{nickel2017poincare, ganea2018hyperbolic, peng2021hyperbolic} excels at representing hierarchical structures due to its natural ability to embed tree-like data. However, applying hyperbolic methods to TKG extrapolation presents several challenges.

\textbf{Dependence on Modulus for Hierarchical Structuring}: Hyperbolic models often rely on modulus-based hierarchy~\cite{nickel2017poincare}, necessitating the omission of normalization techniques during training to preserve modulus information. This results in slow convergence and suboptimal performance in deeper networks~\cite{ioffe2015batch,ba2016layer}, leading to shallow inductive models~\cite{chami2020low,montella2021hyperbolic,han2020dyernie}. Efforts to mitigate this, such as spatial curvature transformations between hyperbolic layers \cite{chami2019hyperbolic}, have been insufficient.

\textbf{Instability in Tangent Space Transformations}: Many hyperbolic methods involve transforming parameters from hyperbolic manifolds to tangent space~\cite{ganea2018hyperbolic,sohn2022bending}, a specialized Euclidean space distinct from the traditional one used in semantic modeling. Without normalization, deep networks in tangent space can produce unstable embedding norms, leading to numerical instability~\cite{nickel2018learning} when mapping back to the manifolds due to limited computational precision, ultimately degrading model performance.

To address the challenges, we propose ETH, a novel hybrid \textbf{E}uclidean-\textbf{T}angent-\textbf{H}yperbolic space model that leverages the strengths of both Euclidean and hyperbolic modeling, resolving limitations of single-space models. Our approach begins by capturing complex semantic information in \textbf{Euclidean} space through a fact co-occurrence and autoregressive method, incorporating normalization for stability. We then transform embeddings into \textbf{Tangent} space with a new scaling mechanism, preserving semantic richness while enabling hierarchical learning through a query-candidate separated modeling approach. These embeddings are subsequently mapped into \textbf{Hyperbolic} space, where hierarchical features are naturally represented. Finally, a hybrid inductive bias for hierarchical and semantic learning is achieved by combining hyperbolic and Euclidean scoring functions, accomplished by a learnable query-specific mixing coefficient that adapts to different query characteristics.

We validate ETH on four TKG benchmark datasets, showing up to a 15.0\% relative error reduction in mean reciprocal rank (MRR) on YAGO compared to existing single-space models. Visualization analysis further confirms the model's adaptability to datasets with varying semantic and hierarchical complexity.

Our contributions can be summarized as follows:
\begin{itemize}
    \item Proposing a multi-space hybrid architecture that integrates hierarchical learning in hyperbolic space with semantic learning in Euclidean space, bridged by tangent space.
    \item Introducing a novel tangent space transformation technique that preserves semantic information while facilitating hierarchical learning.
    \item Developing a hybrid scoring function with a query-specific mixing coefficient, optimizing performance across diverse query types.
\end{itemize}

\section{Related Work}
\subsection{Static KG Reasoning Models}
Static KG reasoning models embed entities and relations into low-dimensional vector spaces to infer missing facts. TransE~\cite{bordes2013translating}, a foundational model, represents relationships as translations between entity embeddings, inspiring various extensions to capture complex relational patterns. Graph Convolutional Networks (GCNs)~\cite{kipf2016semi} have furthered this field, with Relational GCN (RGCN)~\cite{schlichtkrull2018modeling} incorporating relation-specific filters, and Weighted GCN (WGCN)~\cite{shang2019end} introducing learnable relation-specific weights. CompGCN~\cite{vashishth2019composition} enhances link prediction by integrating nodes and relations, while Variational RGCN (VRGCN)~\cite{ye2019vectorized} introduces probabilistic embeddings. Despite their success in static KGs, these models struggle with temporal dynamics and future event prediction.

\subsection{Temporal KG Reasoning Models}
TKG reasoning models extend static approaches by incorporating temporal dynamics to predict future facts. These models operate in two key settings: interpolation and extrapolation. Interpolation~\cite{xu2020temporal} infers missing facts at historical timestamps. Early models like TA-DistMult~\cite{garcia2018learning}, TA-TransE~\cite{garcia2018learning}, and TTransE~\cite{leblay2018deriving} embed temporal information directly into relation embeddings, while HyTE~\cite{dasgupta2018hyte} uses a hyperplane for each timestamp. However, they struggle with predicting future events. Extrapolation predicts future events based on historical data. Know-Evolve~\cite{trivedi2017know} uses temporal point processes, DyREP~\cite{trivedi2019dyrep} models relationship evolution, and RE-NET~\cite{jin2019recurrent} employs sequence-based approaches. More recent methods, like TANGO~\cite{han2021learning} and xERTE~\cite{han2020explainable}, introduce continuous-time reasoning and graph-based reasoning. RE-GCN~\cite{li2021temporal} captures entire KG sequences to enhance efficiency. Despite advancements, most TKG models overlook hierarchical structures, focusing mainly on temporal dynamics, which limits their ability to fully represent the complexity of real-world data.

\subsection{Hyperbolic Models}
Hyperbolic models~\cite{sun2020knowledge} excel at representing hierarchical structures in KGs, often surpassing Euclidean models in this regard. Poincaré embeddings~\cite{nickel2017poincare}, laid the foundation for modeling hierarchies in hyperbolic space. Hyperbolic Neural Networks (HNN)~\cite{ganea2018hyperbolic} further developed this by optimizing embeddings in tangent space. MuRP~\cite{balazevic2019multi} extended these ideas to KGs, refining hyperbolic distances to better capture relational structures. DyERNIE~\cite{han2020dyernie} uses a product manifold for temporal dynamics, while AttH~\cite{chami2020low} employs relation-specific transformations to capture hierarchical levels. HERCULES~\cite{montella2021hyperbolic} adapts AttH to temporal contexts, and HyperVC~\cite{sohn2022bending} brings RE-NET into hyperbolic space, though with moderate success. ReTIN~\cite{jia2023extrapolation} builds on AttH for temporal reasoning by integrating global and real-time embeddings. However, these models often underutilize the semantic strengths of Euclidean space. Our approach addresses this gap by sequentially leveraging Euclidean space for semantic learning and hyperbolic space for hierarchical modeling, integrating the strengths of both geometries.

\section{Problem Formulation and Background}
\subsection{Problem Definition}
In this paper, a TKG is defined as $\mathcal{G}(\mathcal{V},\mathcal{E},\mathcal{T},\mathcal{F})$ , where $\mathcal{V}$, $\mathcal{E}$, and $\mathcal{T}$ represent the sets of entities, relations, and timestamps, respectively, and $\mathcal{F}\subseteq\mathcal{V}\times\mathcal{E}\times\mathcal{V}\times\mathcal{T}$ is the set of all quadruples $(s,r,o,t)$. The TKG can be viewed as a sequence of KG snapshots, denoted by $\mathcal{G}=\{\mathcal{G}_0,\mathcal{G}_1,\cdots,\mathcal{G}_t,\cdots\}$, where each snapshot $\mathcal{G}_t=\{(s,r,o)\mid(s,r,o,t)\in\mathcal{F}\}$ corresponds to a specific timestamp. The TKG extrapolation task aims to predict the set of queries $\mathcal{Q}_{t+1}=\{(q,r)\mid(q,r,o)\in\mathcal{G}_{t+1}\}$, given the most recent $m$ snapshots $\mathcal{G}_{t-m+1:t}\subseteq\mathcal{G}$. Candidates for $\mathcal{Q}_{t+1}$ are drawn from $\mathcal{V}$, denoted by $a\in\mathcal{V}$. The objective is to score each quadruple $(q, r, a, t+1)$ using a scoring function $f_s: \mathcal{V} \times \mathcal{E} \times \mathcal{V} \times \mathcal{T} \rightarrow \mathbb{R}$, where a higher score indicates a greater likelihood that the $a$ is the correct entity. To enhance structural connectivity of the TKG, inverse quadruples $(o, r^{-1}, s, t)$ are also incorporated.

\subsection{Hyperbolic Geometry}
Hyperbolic geometry differs from Euclidean geometry in its parallel postulate, where through any point not on a line, infinitely many lines can be draw parallel to the given line. This leads to exponential growth in the area and perimeter, reflecting the constant negative curvature of hyperbolic space, making it well-suited for modeling hierarchical structures.

The Poincaré ball model is a common representation of hyperbolic space, defined as a $d$-dimential ball $\mathbb{B}_c^d=\{\boldsymbol{x}\in\mathbb{R}^d\mid\Vert \boldsymbol{x}\Vert^2<1/c\}$, where $c$ is the negative curvature ($-c<0$) and $\Vert\cdot\Vert$ is the Euclidean $L_2$ norm. Each point $\boldsymbol{x}\in\mathbb{B}_c^d$ is associated with a tangent space $\mathcal{T}_{\boldsymbol{x}} \mathbb{B}_c^d$, a $d$-dimentional vector space that containing all possible velocity vectors at $\boldsymbol{x}$ on the manifold. 

To transition between the tangent space and the hyperbolic ball, the exponential map $\exp_{\boldsymbol{x}}^c : \mathcal{T}_{\boldsymbol{x}} \mathbb{B}_c^d\rightarrow\mathbb{B}_c^d$, and the logarithmic map $\log_{\boldsymbol{x}}^c : \mathbb{B}_c^d\rightarrow\mathcal{T}_{\boldsymbol{x}} \mathbb{B}_c^d$ are used. Specifically, at the origin $\boldsymbol{0}\in\mathbb{B}_c^d$, these maps are defined as:
\begin{equation}
\begin{aligned}
\exp_{\boldsymbol{0}}^c (\boldsymbol{v})&=\tanh (\sqrt c \Vert \boldsymbol{v}\Vert) \frac{\boldsymbol{v}}{\sqrt c \Vert \boldsymbol{v}\Vert},\\
\log_{\boldsymbol{0}}^c (\boldsymbol{u})&=\operatorname{arctanh} (\sqrt c \Vert \boldsymbol{u}\Vert) \frac{\boldsymbol{u}}{\sqrt c \Vert \boldsymbol{u}\Vert},
\end{aligned}
\end{equation}
where $\boldsymbol{v}\in\mathcal{T}_{\boldsymbol{0}} \mathbb{B}_c^d$ and $\boldsymbol{u}\in\mathbb{B}_c^d$.

The Poincaré geodesic distance between any two points $\boldsymbol{x}$ and $\boldsymbol{y}\in\mathbb{B}_c^d$ is:
\begin{equation}
d^c(\boldsymbol{x},\boldsymbol{y}) = \frac{2}{\sqrt c} \operatorname{arctanh} (\sqrt c \Vert -\boldsymbol{x} \oplus ^c \boldsymbol{y}\Vert),
\end{equation}
where $\oplus ^c$ represents Möbius addition~\cite{ganea2018hyperbolic}, defined as: 
\begin{equation}
\boldsymbol{x}\oplus ^c \boldsymbol{y}
=\frac
{(1+2c\langle \boldsymbol{x},\boldsymbol{y} \rangle + c \Vert \boldsymbol{y}\Vert ^2 )\boldsymbol{x}+(1-c \Vert \boldsymbol{x} \Vert ^2 )\boldsymbol{y}}
{1+2c\langle \boldsymbol{x},\boldsymbol{y} \rangle + c^2 \Vert \boldsymbol{x} \Vert ^2 \Vert \boldsymbol{y}\Vert^2}.
\end{equation}
where $\langle\cdot\rangle$ is Euclidean dot product.

\section{Methodology}
\begin{figure*}[ht]
    \centering
    \includegraphics[width=0.8\textwidth]{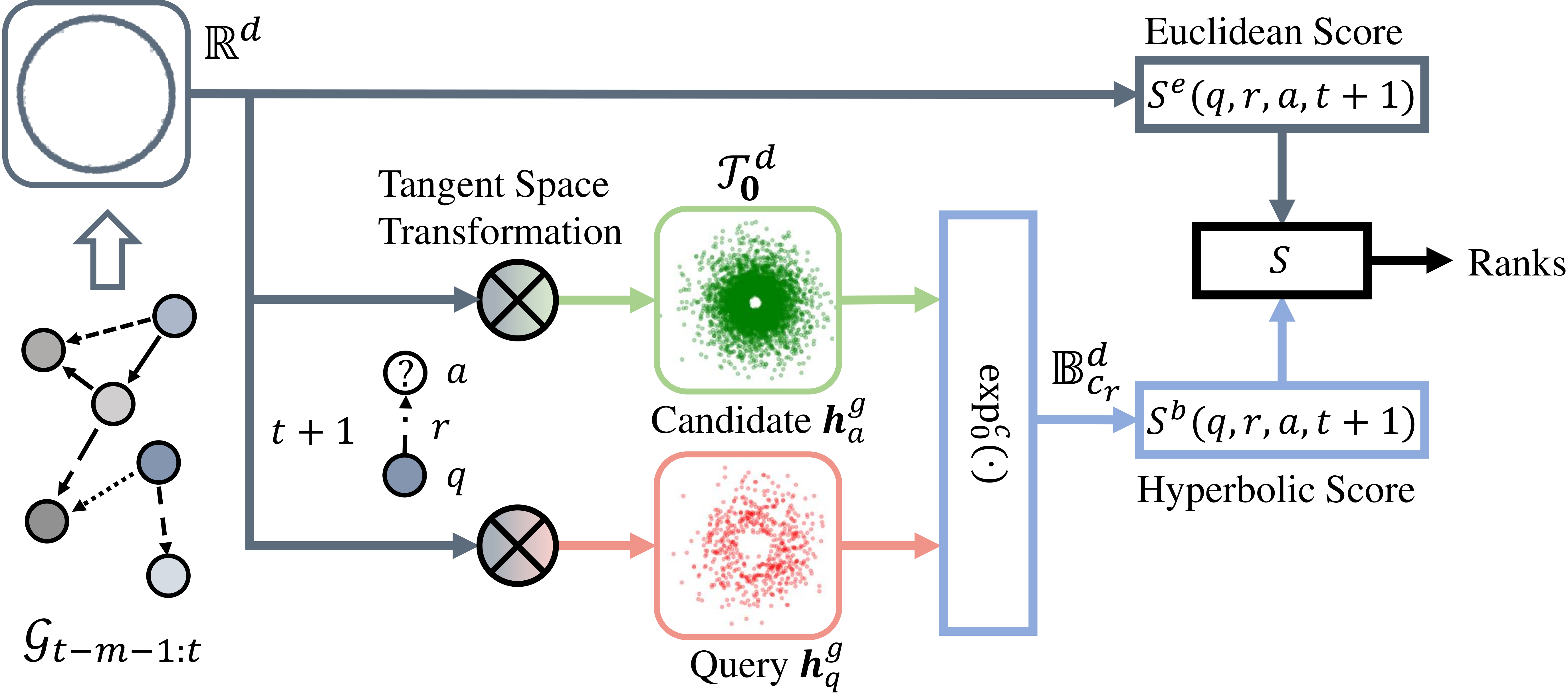}
    \caption{ An illustrative diagram of the proposed ETH model. }
    \label{fig:Archi}
\end{figure*}

This section details ETH, as illustrated in Figure \ref{fig:Archi}. The model first captures complex semantics of multi-relational graphs and dynamic temporal information through a fact co-occurrence and autoregressive method, incorporating normalization throughout (Section~\ref{sec:4.1}). Subsequently, it transforms Euclidean semantic embeddings into tangent space with a new scaling mechanism, preserving semantic information while enabling hierarchical learning. Query and candidate entities are modeled separately to enhance the capture of both semantic and hierarchical information (Section~\ref{sec:4.2}). Finally, the embeddings transition from tangent to hyperbolic space, where a hyperbolic scoring function evaluate quadruples alongside Euclidean scoring on the previously processed vectors. A learnable query-specific scoring coefficient balances semantic and hierarchical modeling for each query (Section~\ref{sec:4.3}). Optimization strategies are discussed in Section~\ref{sec:4.4}.

\subsection{Euclidean Modeling}
\label{sec:4.1}
Semantic information in TKGs arises from graph structures and temporal dynamics. To effectively capture this complexity, entity embeddings are initially encoded in Euclidean space using a Relation-aware Graph Convolutional Network (RGCN) and a Gated Recurrent Unit (GRU). The RGCN captures intra-snapshot graph semantics, while the GRU models temporal dynamics in an autoregressive manner.

The input is a sequence of the last $m$ snapshots $\mathcal{G}_{t-m+1:t}$, used as historical context for predicting queries $\mathcal{Q}_{t+1}$. Entity embeddings $\boldsymbol{h}$ and relation embeddings $\boldsymbol{v}^e\in \mathbb{R}^d$ are initialized randomly, with explicit encoding applied only to entities due to their greater number relative to relations.

\subsubsection{Multi-Relational Graph Semantic Modeling.}
Each snapshot is treated as a multi-relational graph. Entities are encoded based on their connections via a relation-aware GCN, capturing the co-occurrence patterns. For entity $o$ at timestamp $k$ with neighbors $(s,r)\in\mathcal{N}_o^k$, the graph semantic encoding from layer $i$ to $i+1$ in the RGCN with total $l$ layers is given by:
\begin{equation}
\boldsymbol{h}_{k,o}^{i+1}=
f\left(\frac{1}{|\mathcal{N}_o^k |}  \sum_{(s,r)\in\mathcal{N}_o^k} {\boldsymbol{W}_1^i (\boldsymbol{h}_{k,s}^i+\boldsymbol{v}_r^e)} + \boldsymbol{W}_2^i \boldsymbol{h}_{k,o}^i \right),
\label{stage1-1}
\end{equation}
where $\boldsymbol{h}_{k,o}^{i+1}\in\mathbb{R}^d$ is embedding of entity $o$ at layer $(i+1)$, $\boldsymbol{W}_1^i$, $\boldsymbol{W}_2^i \in\mathbb{R} ^{d\times d}$ are learnable weights at layer $i$, and $f(\cdot)$ is the RReLU activation function. Self-loop edges are added for all entities.

\subsubsection{Autoregressive Temporal Semantic Modeling.}
Temporal dynamics are captured using a GRU, which updates semantic embeddings over time:
\begin{equation}
\boldsymbol{h}_k=\operatorname{GRU} (\boldsymbol{h}_{k-1},\boldsymbol{h}_k^l ),
\label{stage1-2}
\end{equation}
where $\boldsymbol{h}_k^l\in\mathbb{R}^d$ is the RGCN output at timestamp $k$. Layer normalization and a scaling factor $\sqrt{d}$ are applied to $\boldsymbol{h}_k$, $\boldsymbol{h}_k^l$, and $\boldsymbol{v}_r^e$ to constrain their $L_2$ norms around $1$. This Euclidean space encoding allows the model to effectively capture complex semantic features early in the processing.

\subsection{Tangent Space Transformation}
\label{sec:4.2}
In Euclidean space modeling, normalization erases hierarchical information. To restore this and prepare for the transition to hyperbolic space, we transform entity embeddings from Euclidean $\mathbb{R}^d$ to tangent space $\mathcal{T}_{\boldsymbol{0}}^d$. This transformation allows for hierarchical relearning, capturing both semantic and hierarchical structures while preventing numerical issues in the Poincaré ball. We employ a dual-mode approach to model distinct behaviors for query entities $q$ and candidate entities $a$, enhancing the model's performance in extrapolation tasks.

\subsubsection{Transformation for Candidate Entity.}
To capture candidates' semantic and hierarchical features, we first apply a linear transformation to the entity embedding $\boldsymbol{h}_t$:
\begin{equation}
\boldsymbol{h}_a^e=\boldsymbol{W}_1^e \boldsymbol{h}_t+\boldsymbol{b}_1^e,
\label{stage2-1}
\end{equation}
where $\boldsymbol{W}_1^e \in \mathbb{R}^{d \times d}$ and $\boldsymbol{b}_1^e \in \mathbb{R}^d$ are learnable parameters. This transformation captures the necessary features before transitioning to tangent space, where the core transformation is:
\begin{equation}
\boldsymbol{h}_a^g=\boldsymbol{W}_1^g \gamma(\boldsymbol{W}^g (\tanh (\boldsymbol{h}_a^e) \otimes \boldsymbol{h}_{t,a})),
\label{stage2-2}
\end{equation}
where $\boldsymbol{W}^g$, $\boldsymbol{W}_1^g\in \mathcal{T}_{\boldsymbol{0}}^{d\times d}$ are weight matrices, $\otimes$ denotes Hadamard product, and $\gamma$ is an optional activation function. The superscript $g$ indicates parameters in tangent space. The tanh function maps $\boldsymbol{h}_a^e$ elements to $[-1,1]$, facilitating stable hierarchical information capture.

\subsubsection{Transformation for Query Entity.}
Query entity embeddings undergo a similar linear transformation, with a slight variation:
\begin{equation}
\boldsymbol{h}_q^e=\boldsymbol{W}_2^e \operatorname{cat} ([\boldsymbol{h}_{t,q};\boldsymbol{v}_r^e ]\mid (q,r)\in\mathcal{Q}_{t+1} )+\boldsymbol{b}_2^e,
\label{stage2-3}
\end{equation}
where $\operatorname{cat}(\cdot)$ concatenates the query entity embedding $\boldsymbol{h}_{t,q}$ and relation embedding $\boldsymbol{v}_r^e$. Here, $\boldsymbol{W}_2^e \in \mathbb{R}^{2d \times d}$ and $\boldsymbol{b}_2^e \in \mathbb{R}^d$ are the parameters used to model these concatenated embeddings. The subsequent tangent space transformation is defined as:
\begin{equation}
\boldsymbol{h}_q^g=\boldsymbol{W}_2^g \gamma(\boldsymbol{W}^g (\tanh(\boldsymbol{h}_q^e )\otimes \boldsymbol{h}_{t,q} )),
\label{stage2-4}
\end{equation}
where $\boldsymbol{W}_2^g\in\mathcal{T}_{\boldsymbol{0}}^{d\times d}$. The shared weight matrix $\boldsymbol{W}^g$ maintains consistency between candidate and query entity embeddings. After these transformations, $\boldsymbol{h}_a^g$ and $\boldsymbol{h}_q^g$ are enriched with hierarchical information and are numerically stable, ready for hyperbolic space modeling.

\subsection{Hyperbolic-Euclidean Hybrid Scoring Function}
\label{sec:4.3}
In the final stage, we integrate semantic and hierarchical modeling through a hybrid scoring function, balancing each query's need for these aspects via a query-specific score mixing approach.

\subsubsection{Euclidean Dot Product Scoring Function.}
The Euclidean dot product scoring function measures semantic similarity between query and candidate embeddings:
\begin{equation}
S^e (q,r,a,t+1)=\langle \boldsymbol{h}_q^e, \boldsymbol{h}_a^e \rangle,
\label{stage3-1}
\end{equation}
where $S^e$ represents the Euclidean dot product score.

\subsubsection{Hyperbolic Distance Scoring Function.}
To capture hierarchical structures, we apply a relation-specific curvature $c_r$ to map embeddings from tangent space to hyperbolic space via exponential transformation:
\begin{equation}
\begin{aligned}
\boldsymbol{h}_a^b&=\exp_{\boldsymbol{0}}^{c_r}(\boldsymbol{h}_a^g ), \\
\boldsymbol{h}_q^b&=\exp_{\boldsymbol{0}}^{c_r}(\boldsymbol{h}_q^g ),
\end{aligned}
\label{stage3-2}
\end{equation}
where $\boldsymbol{h}_a^b$ and $\boldsymbol{h}_q^b\in\mathbb{B}_{c_r}^d$ are embeddings of candidate $a$ and query $q$ in the Poincaré ball. The hyperbolic distance scoring function is then used:
\begin{equation}
S^b (q,r,a,t+1)=-d^{c_r}(\boldsymbol{h}_q^b \oplus ^{c_r} \boldsymbol{v}_r^b, \boldsymbol{h}_a^b )^2+b_q+b_a,
\end{equation}
where $S^b$ represents the hyperbolic distance score, $\boldsymbol{v}_r^b\in\mathbb{B}_{c_r}^d$ is the learnable relation embedding in hyperbolic space, and $b_q$, $b_a\in \mathbb{R}$ are entity-specific biases. Unlike entity embeddings, $\boldsymbol{v}_r^b$ and $\boldsymbol{v}_r^e$ are learned directly in their respective spaces without explicit transformation. This approach ensures that $\boldsymbol{h}_q^b$ is properly adjusted to capture its distinct interaction with relation $r$ compared to its Euclidean counterpart $\boldsymbol{h}_q^e$.

\subsubsection{Hybrid Space Scoring Function.}
We combine the Euclidean and hyperbolic scores using a query-specific mixing coefficient:
\begin{equation}
S(q,r,a,t+1)=\sigma(\beta_{q,r} S^b+(1-\beta_{q,r} ) S^e ),
\end{equation}
where $\sigma(\cdot)$ is the sigmoid function. The coefficient $\beta_{q,r}$ is defined as:
\begin{equation}
\beta_{q,r}=\sigma\left(\frac {\langle \boldsymbol{s}_q, \boldsymbol{s}_r \rangle} {w}\right),
\label{stage3-3}
\end{equation}
where $\boldsymbol{s}_q, \boldsymbol{s}_r\in\mathbb{R}^w$ are query entity and relation vectors, respectively. The dot product of these vectors, processed through the sigmoid function, ensures that $\beta_{q,r}$ ranges between $0$ and $1$. This approach enables information sharing among queries with common entities or relations.

\begin{table*}[h]
    \centering
    \fontsize{10}{12}\selectfont
    \makebox[\textwidth][c] 
    {
    \begin{tabular*}{0.7\textwidth}{@{\extracolsep{\fill}} lrrrrrrr}
    \hline 
    Datasets & $|\mathcal{V}|$ & $|\mathcal{E}|$  & $|\mathcal{F}_{train}|$ & $|\mathcal{F}_{valid}|$ & $|\mathcal{F}_{test}|$ & $|\mathcal{T}|$ & Time interval \\
    \hline
    ICEWS14 & 6,869 & 230 & 74,845 & 8,514 & 7,371 & 365 & 24 hours \\
    
    ICEWS05-15 & 10,094 & 251 & 368,868 & 46,302 & 46,159 & 4,017 & 24 hours \\
    
    YAGO & 10,623 & 10 & 161,540 & 19,523 & 20,026 & 189 & 1 year \\
    
    WIKI & 12,554 & 24 & 539,286 & 67,538 & 63,110 & 232 & 1 year \\
    \hline
    \end{tabular*}
    }
    \caption{ \label{datasets} Dataset summaries.}
\end{table*}

\subsection{Optimization}
\label{sec:4.4}
We optimize the model by minimizing the cross-entropy loss function:
\begin{equation}
\mathcal{L}=
\sum_{t=0}^{|\mathcal{T}|-1} \sum_{(q,r)}^{Q_{t+1}} \sum_a^\mathcal{V} y_{t+1}^{q,r,a} \log S(q,r,a,t+1),
\end{equation}
where $y_{t+1}^{q,r,a}\in\mathbb{R}$ represents the label for candidate $a$ in query $(q,r)$ at timestamp $t+1$. Most parameters in our model are either in Euclidean space or tangent space, avoiding the complexities of Riemann optimization, thereby enhancing stability and performance.

\section{Experiments}
\subsection{Experiments Setup}
\subsubsection{Datasets.}
\label{sec:dataset}
ETH is evaluated on four widely adopted TKG datasets: ICEWS14, ICEWS05-15~\cite{garcia2018learning}, WIKI~\cite{leblay2018deriving}, and YAGO~\cite{mahdisoltani2013yago3}. For ICEWS14 and ICEWS05-15, we follow standard practice by splitting the datasets into 80\% training, 10\% validation, and 10\% test sets, ensuring chronological order ($t_{\text{train}} < t_{\text{valid}} < t_{\text{test}}$)\cite{jin2019recurrent}. Dataset details are summarized in Table~\ref{datasets}.

\subsubsection{Evaluation Metrics.}
Mean Reciprocal Rank (MRR) and Hits@1/3/10 are used as evaluation metrics. Among the various metric settings: raw~\cite{bordes2011learning}, static filter~\cite{bordes2013translating}, and time filter~\cite{han2020graph}. Time filter is preferred according to~\cite{gastinger2022evaluation} for extrapolation tasks, which excludes other correct answers from the ranking process when a query $(q, r, ?, t)$ has multiple correct answers at the same timestamp. This approach is justified as the other answers are equally valid. Hence, we report results exclusively under the time filter setting.

\subsubsection{Implementation Details.}
Embedding dimensions $d$ and $w$ are set to 200, with the RGCN layer count $l$ at 2 for ICEWS14 and ICEWS05-15, and 1 for WIKI and YAGO. A grid search within the range [1, 30] determined optimal history lengths $m$ as 10, 24, 2, and 2 for ICEWS14, ICEWS05-15, YAGO, and WIKI, respectively. The activation function $\gamma$ is set to ReLU for ICEWS14, YAGO, and WIKI, and None for ICEWS05-15. Adam optimizer is used with a 0.001 learning rate. Training was conducted on a GeForce RTX 4060 TI GPU. For comparisons with static methods, timestamps were excluded during training and testing.

\subsubsection{Compared Mothods.}
ETH is compared against baseline hyperbolic models AttH~\cite{chami2020low} and HERCULES~\cite{montella2021hyperbolic}, as well as Euclidean models RGCRN~\cite{seo2018structured}, RE-NET~\cite{jin2019recurrent}, CyGNet~\cite{zhu2021learning}, xERTE~\cite{han2020explainable}, TLogic~\cite{liu2022tlogic}, and EvoKG~\cite{park2022evokg} for TKG extrapolation tasks. Hyperbolic baseline results are from~\cite{jia2023extrapolation}, and Euclidean baseline results from~\cite{liang2024survey}.

\begin{figure}[htb]
    \centering
    \includegraphics[width=1\columnwidth]{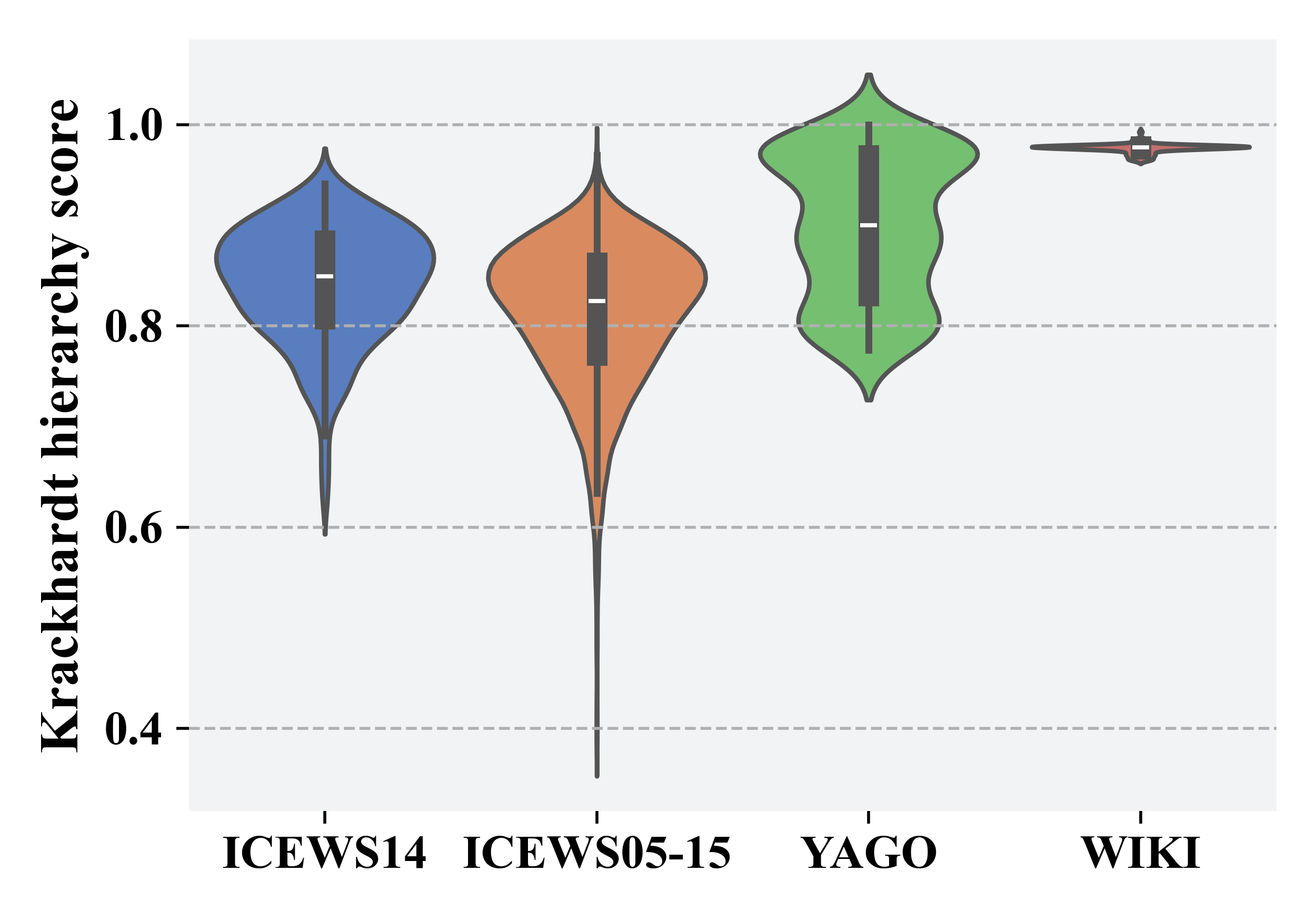}
    \caption{ $Khs$ statistics for each dataset. }
    \label{fig:Khs}
\end{figure}

\begin{table*}[htb]
\centering
\fontsize{9}{12}\selectfont
\makebox[\textwidth][c] 
{
\setlength{\tabcolsep}{1mm}
\begin{tabular*}{1\textwidth}{@{\extracolsep{\fill}} ccccccccccccccccccccc}
\hline
\multicolumn{1}{c}{\multirow{2}{*}{Model}} & & \multicolumn{4}{c}{ICE14} & & \multicolumn{4}{c}{ICE05-15} & & \multicolumn{4}{c}{YAGO} & & \multicolumn{4}{c}{WIKI} \\
\cline{3-6} \cline{8-11} \cline{13-16} \cline{18-21} \multicolumn{1}{c}{} & &
MRR & H@1 & H@3 & H@10 & & MRR & H@1 & H@3 & H@10 & & MRR & H@1 & H@3 & H@10 & & MRR & H@1 & H@3 & H@10 \\ 
\hline
RGCRN &  & 38.48  & 28.52  & 42.85  & 58.10  &  & 44.56  & 34.16  & 50.06  & 64.51  &  & 65.76  & 62.52  & 67.56  & 71.69  &  & 65.79  & 61.66  & 68.17  & 72.99 \\

AttH  &  & 36.10  & 26.30  & 40.10  & 55.60  &  & 36.90  & 26.80  & 41.20  & 56.80  &  & - & - & - & - &  & - & - & - & - \\

RE-NET &  & 39.86  & 30.11  & 44.02  & 58.21  &  & 43.67  & 33.55  & 48.83  & 62.72  &  & 66.93  & 58.59  & 71.48  & 86.84  &  & 58.32  & 50.01  & 61.23  & 73.57 \\

HERCULES  &  & 35.80  & 26.20  & 39.20  & 55.80  &  & 36.80  & 26.50  & 41.20  & 57.00  &  & - & - & - & - &  & - & - & - & - \\

CyGNet &  & 37.65  & 27.43  & 42.63  & 57.90  &  & 40.42  & 29.44  & 46.06  & 61.60  &  & 68.98  & 58.97  & 76.80  & 86.98  &  & 58.78  & 47.89  & 66.44  & 78.70 \\

TANGO &  & - & - & - & - &  & 42.86  & 32.72  & 48.14  & 62.34  &  & 63.34  & 60.04  & 65.19  & 68.79  &  & 53.04  & 51.52  & 53.84  & 55.46 \\

xERTE &  & 40.79  & \textbf{32.70}  & 45.67  & 57.30  &  & 46.62  & \textbf{37.84}  & 52.31  & 63.92  &  & \underline{84.19}  & \underline{80.09}  & \underline{88.02}  & \underline{89.78}  &  & 73.60  & 69.05  & 78.03  & 79.73 \\

RE-GCN &  & \underline{42.00}  & 31.63  & 47.20  & \underline{61.65}  &  & \underline{48.03}  & 37.33  & \underline{53.90}  & \underline{68.51}  &  & 82.30  & 78.83  & 84.27  & 88.58  &  & \underline{78.53}  & \underline{74.50}  & \underline{81.59}  & 84.70 \\

TLogic &  & 41.80  & 31.93  & \underline{47.23}  & 60.53  &  & 45.99  & 34.49  & 52.89  & 67.39  &  & - & - & - & - &  & - & - & - & - \\

EvoKG &  & 27.18  & - & 30.84  & 47.67  &  & - & - & - & - &  & 68.59  & - & 81.13  & 92.73  &  & 68.03  & - & 79.60  & \textbf{85.91} \\

Our Model &  & \textbf{42.68}  & \underline{32.19}  & \textbf{47.86}  & \textbf{62.88}  &  & \textbf{48.38}  & \underline{37.64}  & \textbf{54.18}  & \textbf{68.92}  &  & \textbf{86.56}  & \textbf{83.33}  & \textbf{89.00}  & \textbf{91.67}  &  & \textbf{80.34}  & \textbf{76.62}  & \textbf{83.55}  & \underline{85.81} \\ 
\hline
\end{tabular*}
}
\caption{ \label{comp1} Performance (\%) on extrapolation tasks for ICEWS14, ICEWS05-15, YAGO, and WIKI under the time filter setting. Best scores are in \textbf{bold}, second-best are \underline{underlined}.}
\end{table*}

\subsection{Performance Comparison}
Table~\ref{comp1} presents the extrapolation task results, showcasing ETH's effectiveness across four datasets. ETH consistently outperforms baseline models, demonstrating superior ability to capture both semantic and hierarchical information. Notably, ETH surpasses hyperbolic models such as AttH and HERCULES by effectively capturing semantic nuances in Euclidean space and outperforms Euclidean models. We calculate Krackhardt hierarchy scores ($Khs$)\cite{krackhardt2014graph} for every snapshot in each dataset, with statistics shown in Figure~\ref{fig:Khs}. Higher $Khs$ indicate a more hierarchical, tree-like structure, where hyperbolic embeddings perform particularly well, as seen in datasets like YAGO and WIKI. Specifically, ETH achieves relative error reductions on the YAGO, with 15.00\% in MRR, 16.27\% in Hits@1, 8.18\% in Hits@3, and 18.49\% in Hits@10, compared to the second-best model. On the WIKI dataset, ETH records relative error reductions of 8.43\% in MRR, 8.31\% in Hits@1, and 10.64\% in Hits@3. 

ETH's strong performance on YAGO and WIKI, both characterized by significant time intervals and pronounced hierarchy, illustrates its effective use of hierarchical information via tangent space transformation. While ETH trails xERTE in Hits@1 on the ICEWS14 and ICEWS05-15 datasets, it still leads in MRR and Hits@3/10, indicating that the hybrid scoring mechanism captures a more comprehensive range of semantic and hierarchical information, which leads to robust predictions. Despite RE-GCN being a strong Euclidean competitor, ETH consistently outperforms it, particularly on YAGO and WIKI, underscoring the importance of hierarchical information in temporal knowledge graph reasoning.

\subsection{Ablation Studies}
To evaluate the contribution of each component within ETH, ablation studies are conducted, as summarized in Table~\ref{Ablation}.

\subsubsection{Impact of Euclidean Semantic Modeling.}
The importance of Euclidean semantic modeling (Equations~\ref{stage1-1} and~\ref{stage1-2}) is assessed by removing this component, retaining only the Tangent and Hyperbolic spaces with randomly initialized embeddings (denoted as -se). The results reveal a significant performance drop across all datasets, underscoring the critical role of semantic information in TKG extrapolation.

\subsubsection{Impact of Tangent-Hyperbolic Hierarchical Modeling.}
The Tangent-Hyperbolic modeling is examined through two experiments: -tst and -q. In the -tst configuration, the tangent space transformation (Equations~\ref{stage2-2} and~\ref{stage2-4}) is removed, with $\boldsymbol{h}_t$ directly fed into the hyperbolic scoring function. This leads to performance declines, especially on YAGO and WIKI, emphasizing the importance of hierarchical relearning. In the -q setup, query embedding modeling (Equations~\ref{stage2-3} and~\ref{stage2-4}) is replaced with basic vector addition $\boldsymbol{h}_q + \boldsymbol{v}_r^e$. This modification results in substantial performance losses, confirming the necessity of distinct modeling for query and candidate entities.

\begin{table}[t]
    \centering
    \fontsize{9}{12}\selectfont
    \setlength{\tabcolsep}{3mm}
    \begin{tabular}{lcccc}
    \hline
    Model & ICE14 & ICE05-15 & YAGO & WIKI  \\ 
    \hline
    Our Model & \textbf{42.68}  & \textbf{48.38}  & \textbf{86.56}  & \textbf{80.34} \\
    -se & 38.53  & 38.74  & 59.24  & 48.66 \\
    -tst & 42.59  & 48.30  & 82.76  & 79.50 \\
    -q & 35.34  & 33.58  & 75.47  & 76.44 \\
    $\beta_{q,r}=0$ & 19.49  & 11.51  & 72.45  & 76.35 \\
    $\beta_{q,r}=1$ & 40.46  & 46.50  & 76.07  & 74.98 \\
    $\beta_{q,r}$ learned & 41.63  & 47.67  & 82.66  & 77.83 \\
    \hline
    \end{tabular}
    \centering \caption{ \label{Ablation} MRR (\%) for ablation studies.}
\end{table}

\begin{figure}[t]
    \centering
    \subfigure[ICEWS14]{
        \centering
        \includegraphics[width=0.45\columnwidth]{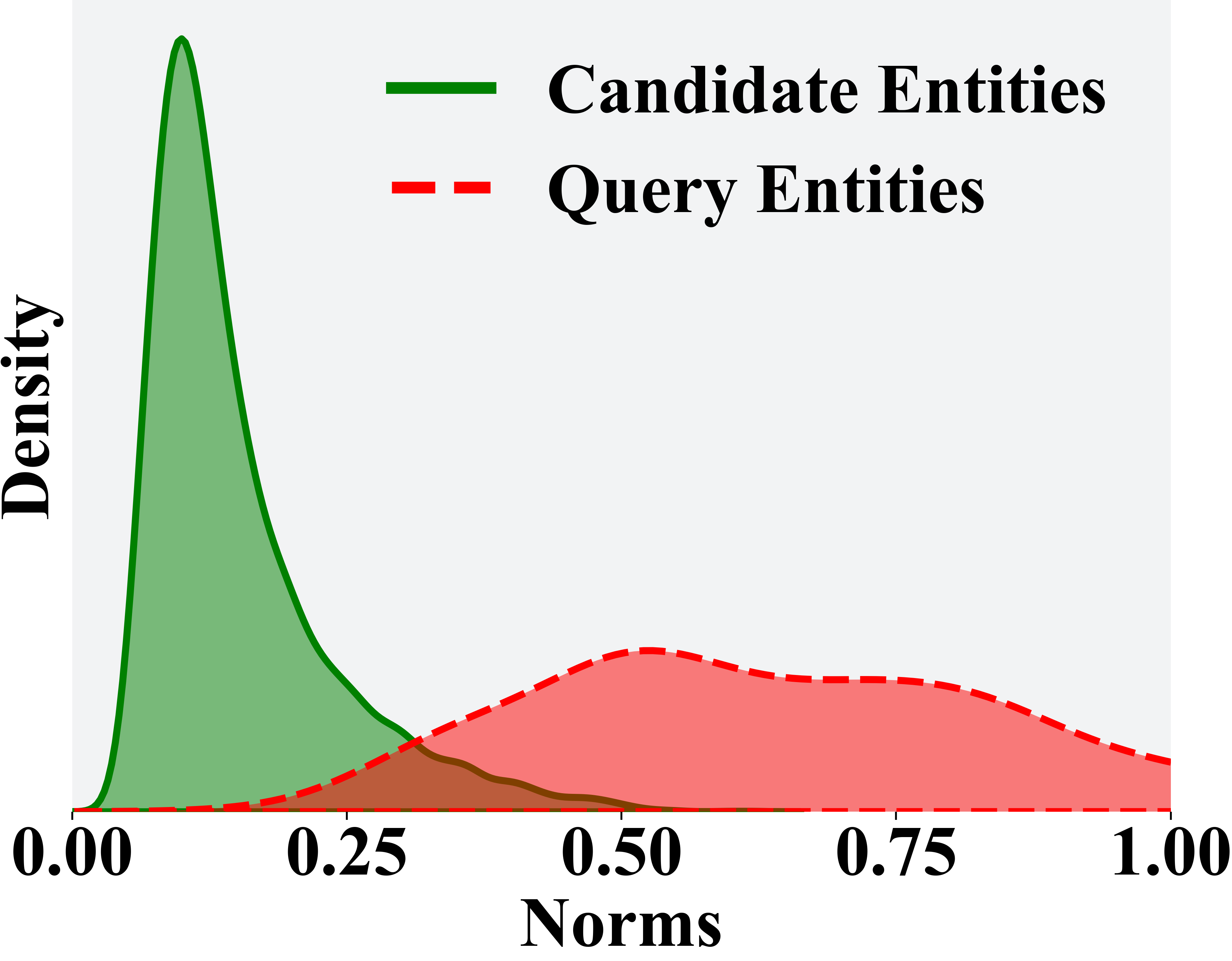}
    }
    \hfill 
    \subfigure[YAGO]{
        \centering\
        \includegraphics[width=0.45\columnwidth]{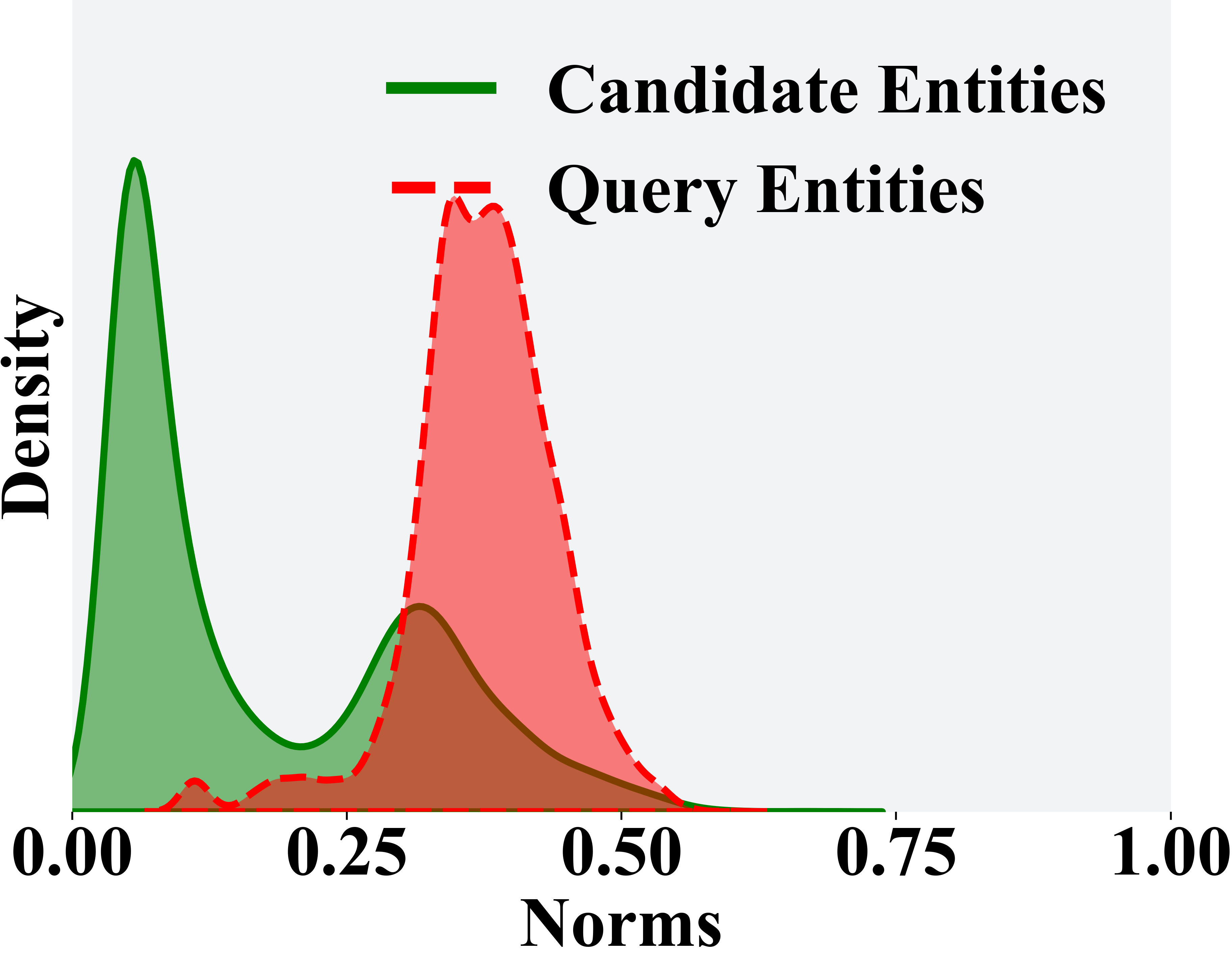}
    }
    \caption{\label{fig:norms} Density distribution of $L_2$ norms in Tangent space for candidate and query entities in ICEWS14~a and YAGO~b.}
\end{figure}

\subsubsection{Impact of Hybrid Scoring.}
Hybrid scoring is evaluated by setting $\beta_{q,r}$ to 0, 1, and allowing it to be learned directly (Table~\ref{Ablation}). Setting $\beta_{q,r} = 0$ disables the contribution of the tangent space, causing significant performance drops on ICEWS14 and ICEWS05-15, likely due to gradient vanishing in longer history settings. In contrast, YAGO and WIKI, with shorter histories, do not exhibit this issue, highlighting hybrid scoring's role in preventing gradient vanishing and enhancing robustness. Setting $\beta_{q,r} = 1$ confines the model to hyperbolic distance scoring, which, while stable, is less effective than the hybrid approach. Allowing $\beta_{q,r}$ to be learned directly underperforms compared to the inductive approach, where entity-relation interactions drive $\beta_{q,r}$, conveying richer information for superior performance.

\subsection{Visualization Analysis}
\subsubsection{Tangent Transformation Analysis.}
Figure~\ref{fig:norms} illustrates the density distributions of $L_2$ norms for candidate and query entities in Tangent space ($\Vert h^g_a\Vert$ and $\Vert h^g_q\Vert$) for the ICEWS14 (Figure~\ref{fig:norms}a) and YAGO (Figure~\ref{fig:norms}b) datasets. The embeddings in Tangent space appear stretched and scaled down, contributing to the model's robustness when $\beta_{q,r}=1$, as it mitigates gradient vanishing issues. Additionally, query embeddings exhibit larger, more varied norm distributions compared to candidates, indicating greater diversity in hierarchical and semantic features. This aligns with the observed performance drop when query modeling is omitted. Furthermore, the YAGO dataset shows multiple peaks in norm distributions, unlike the single peak in ICEWS14, reflecting YAGO's more diverse and hierarchical structure, as supported by the $Khs$ distribution in Figure~\ref{fig:Khs}. This adaptability underscores the model's robustness.

\begin{figure}[t]
    \centering
    \includegraphics[width=1\columnwidth]{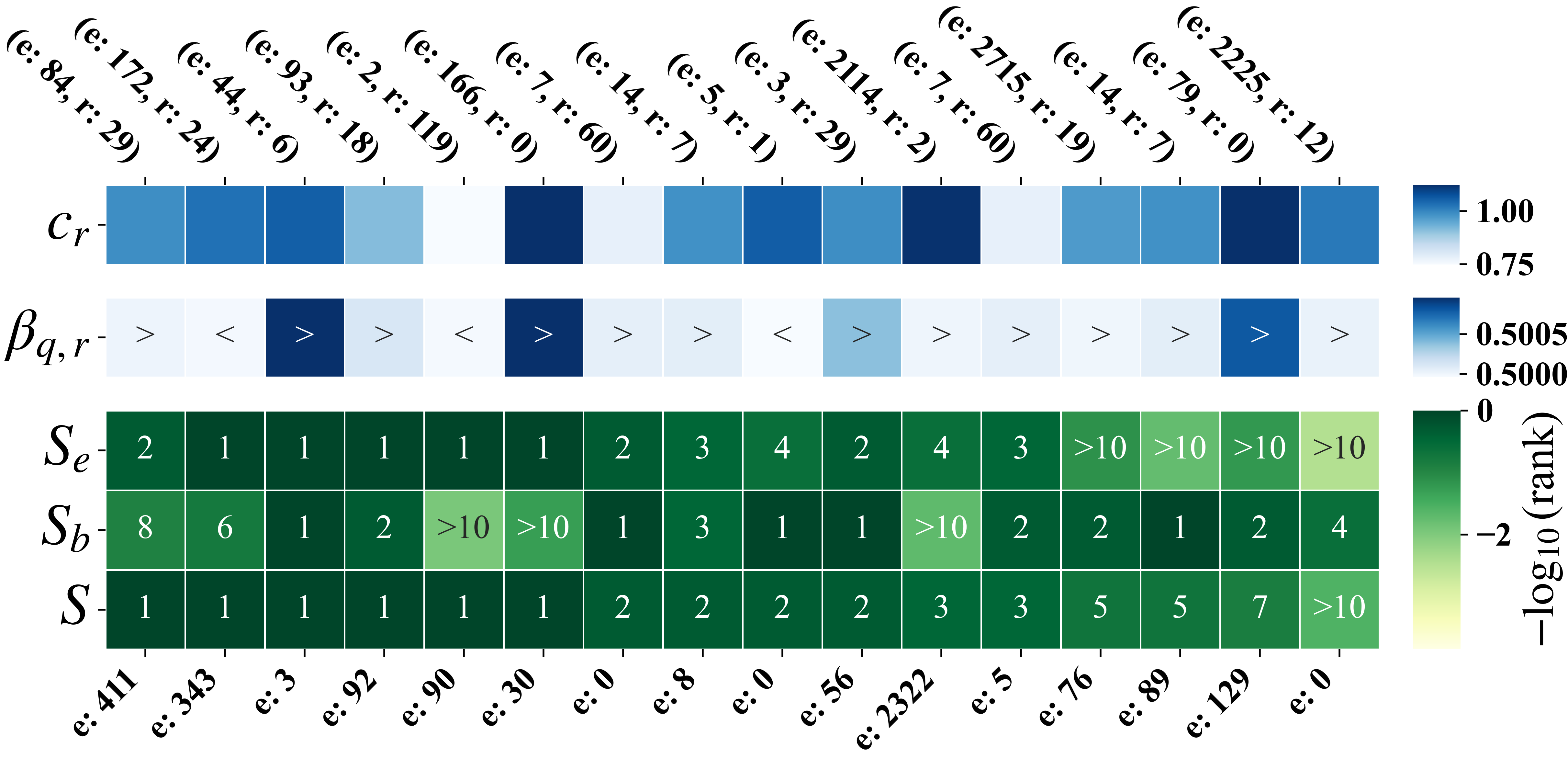}
    % \makebox[\columnwidth]{\includegraphics[width=1.1\columnwidth]{images-ICEWS14s-rank_smaples.png}}
    \caption{ICEWS14 scoring examples.}
    \label{fig:ice14-examples}
\end{figure}

\subsubsection{Hybrid Scoring Analysis.}
Figure~\ref{fig:ice14-examples} shows randomly selected scoring examples from the testing phase on ICEWS14. The top ticks indicate query IDs (entity and relation), while the bottom ticks represent the correct entity IDs. In the $c_r$ heatmap, color intensity reflects absolute values; in the $\beta_{q,r}$ heatmap, color represents the value, with "$<$" and "$>$" indicating values below or above 0.5. The final heatmap shows the rank of the correct entity, with colors processed as $-\log_{10}($rank$)$ and annotations indicating the actual rank. The figure demonstrates the model's ability to adjust $c_r$ for each relation and $\beta_{q,r}$ for each query, showing effective collaboration between Euclidean and Hyperbolic scores for more accurate rankings.

\section{Conclusions}
This paper presents ETH, a hybrid model that integrates Euclidean and hyperbolic spaces, bridged through tangent space, for temporal knowledge graph reasoning. By employing multi-space modeling, ETH effectively captures both semantic and hierarchical information. The model transitions embeddings from Euclidean space, through tangent space, into hyperbolic space, preserving semantic integrity while enhancing hierarchical learning. Experimental results demonstrate ETH's superiority over single-space models, with visualization analyses confirming its adaptability across diverse datasets. Future directions for this work include exploring other tasks that could benefit from the hybrid geometric space framework. Additionally, the proposed tangent space transformation can also be extended to other hyperbolic methods. 

\bibliography{aaai25}

\end{document}